\documentclass{article}
\usepackage{spconf,amsmath,graphicx,hyperref}
\usepackage{multirow}
\usepackage{booktabs}
\usepackage{amssymb}
\usepackage{xcolor}

\title{Collaboratively Guided Adversarial Robust Distillation with Teacher-Favorable Examples}
\name{
Zhi Li$^{1}$,
Haowei Liu$^{1}$,
Hongchen Yang$^{1}$,
Xiaoxuan Wang$^{2}$,
Song Gao$^{1}$\sthanks{Corresponding author: Song Gao (gaos@ynu.edu.cn).},
Shaowen Yao$^{1}$,
Wei Zhou$^{1}$
}

\address{
$^{1}$Engineering Research Center of Cyberspace and School of Software and AI,\\
Yunnan University, Kunming, China\\
$^{2}$School of Information Science and Technology, Yunnan Normal University, Kunming, China\\
}
\begin{document}
\topmargin=0mm

\maketitle

\begin{abstract}
Adversarial distillation transfers robustness from high-capacity teachers to compact students. 
Existing adversarial distillation methods mainly use teacher predictions on clean or adversarial examples to supervise student learning.
However, teacher-favorable supervision within the perturbation neighborhood remains underexplored in adversarial distillation.
We therefore propose Collaboratively Guided Adversarial Robust Distillation (CGARD), which jointly optimizes distinct student-adversarial and teacher-collaborative examples within the same perturbation neighborhood. The teacher-collaborative example is constrained to incur no greater cross-entropy loss under the teacher than the clean input.
CGARD combines collaborative teacher guidance with adversarial teacher supervision to improve robust knowledge transfer.
Experiments on CIFAR-10 and CIFAR-100, including white-box evaluation and additional black-box transfer evaluation, demonstrate consistent robustness improvements over strong adversarial distillation baselines.

\end{abstract}


\section{Introduction}
\label{Introduction}

Lightweight networks enable deployment on platforms with limited resources, but remain vulnerable to adversarial perturbations~\cite{att_CUI2026132185,att_aaicml20,att_deepfool_cvpr2016}.
Adversarial training (AT) improves robustness by incorporating adversarial examples into model optimization~\cite{atpgd_ICLRmadry2018,at_uiat_cvpr2023,AWP_NEURIPS2020}, while several methods further improve the trade-off between clean and robust accuracy~\cite{at_tradesicml2019,AR-AT_ICLR2025,LBGAT_2021_ICCV}. 
However, the limited capacity of lightweight models can constrain the effectiveness of AT, motivating adversarial distillation (AD) to transfer robustness from high-capacity robust teachers to compact students~\cite{ad_mtard_eccv2022}.

\begin{figure*}[t]
\centering
\includegraphics[width=0.9\textwidth]{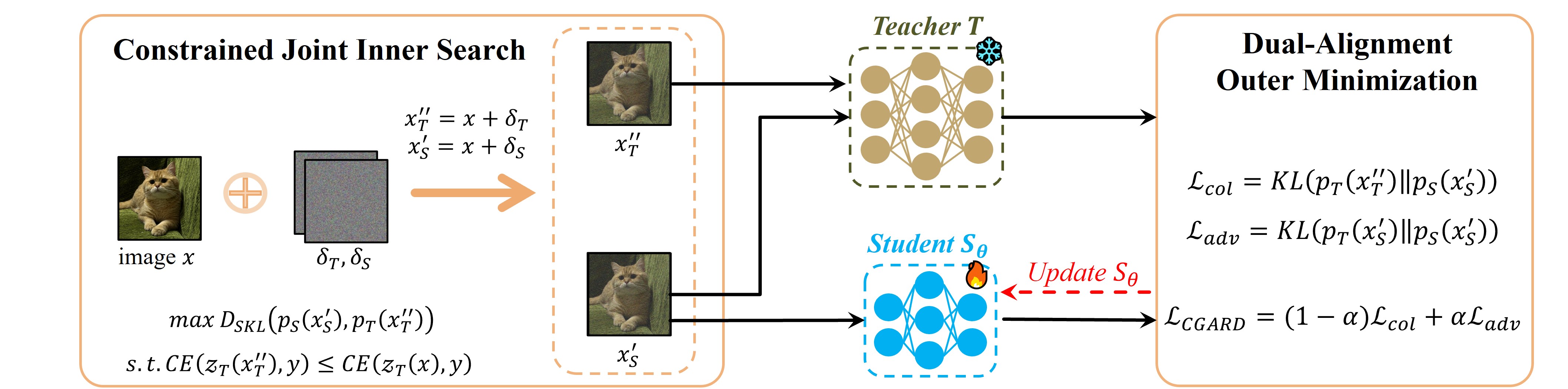}
\caption{Overview of CGARD.
The joint inner search optimizes student-adversarial and teacher-collaborative examples through the discrepancy between teacher and student predictions, while dual alignment uses teacher supervision from both examples.}
\label{fig:framework}
\end{figure*}

Adversarial distillation transfers robustness from a robust teacher to a compact student through teacher supervision~\cite{AD_ard2020aaai}.
RSLAD~\cite{AD_rslad2021iccv} exploits robust soft labels, while IAD~\cite{AD_iad2022iclr} uses student introspection to regulate potentially unreliable teacher guidance.
AdaAD~\cite{AdaAD2023cpvr} adaptively searches a shared perturbed example by maximizing the prediction discrepancy between the teacher and student predictions.
PeerAiD~\cite{AD_peeraid2024cvpr} introduces a specialized peer tutor to provide robustness supervision to the student.

Across these approaches, teacher supervision is usually obtained from the clean input or from an adversarial input generated to challenge the student. Consequently, teacher-favorable inputs within the same perturbation neighborhood remain largely unused. ST~\cite{stat2023iclr} provides a useful precedent in adversarial training by jointly constructing adversarial and low-loss collaborative examples with a single model. Applying this principle to adversarial distillation, however, is nontrivial because it involves a trainable student and a fixed robust teacher. A straightforward adaptation would first construct a teacher-collaborative example and then search for a student-adversarial example using the resulting fixed teacher prediction. Because the teacher-collaborative example remains fixed during the second search, this sequential strategy cannot adapt it to the evolving student-adversarial prediction. We therefore propose Collaboratively Guided Adversarial Robust Distillation (CGARD), as illustrated in Fig.~\ref{fig:framework}. CGARD jointly optimizes the student-adversarial and teacher-collaborative examples as separate variables within the same perturbation neighborhood. The joint search maximizes the discrepancy between the student prediction on the adversarial example and the teacher prediction on the collaborative example. Meanwhile, the teacher-collaborative example is constrained to incur no greater cross-entropy loss under the teacher than the clean input. A dual-alignment objective then supervises the student's adversarial prediction using teacher predictions on both examples. This use of separate search variables distinguishes CGARD from AdaAD~\cite{AdaAD2023cpvr}, which maximizes teacher-student discrepancy on a single perturbed input shared by both models.

Our main contributions are:
\begin{itemize}
\item We introduce a constrained example construction for adversarial distillation, jointly optimizing student-adversarial and teacher-collaborative examples through the discrepancy between teacher and student predictions.
\item We design a dual-alignment objective to combine teacher predictions on the teacher-collaborative and student-adversarial examples for robust knowledge transfer.
\item Experiments on CIFAR-10 and CIFAR-100 demonstrate consistent robustness improvements over strong adversarial distillation baselines, including up to 0.72 percentage points higher AutoAttack accuracy.
\end{itemize}

\section{Methodology}
\label{sec:method}

\subsection{Motivation for Teacher-Collaborative Examples}
\label{subsec:lowce_robustness}

We motivate teacher-collaborative examples through the relationship between teacher cross-entropy and the logit margin.
Let $z_T(x)=[z_{T,1}(x),\ldots,z_{T,C}(x)]\in\mathbb{R}^{C}$ denote the logits of the fixed robust teacher.
For a correctly classified example $(x,y)$, we define the logit margin as
\begin{equation}
\gamma(x)
=
z_{T,y}(x)-\max_{c\neq y}z_{T,c}(x).
\end{equation}

The cross-entropy loss satisfies

\begin{equation}
\begin{aligned}
\mathrm{CE}(z_T(x),y)
&=
\log\left(
1+\sum_{c\neq y}
e^{z_{T,c}(x)-z_{T,y}(x)}
\right) \\
&\geq
\log\left(1+e^{-\gamma(x)}\right),
\end{aligned}
\end{equation}
which implies

\begin{equation}
\gamma(x)
\geq
-\log\left(
e^{\mathrm{CE}(z_T(x),y)}-1
\right).
\label{eq:ce_margin}
\end{equation}

We further relate the logit margin to the perturbation required to change the teacher prediction.
Assume that the teacher logit mapping is $L$-Lipschitz under $\ell_\infty$:

\begin{equation}
\|z_T(x)-z_T(x')\|_\infty
\leq
L\|x-x'\|_\infty .
\end{equation}

For any perturbation $\delta$ that changes the predicted label, there exists a class
$c\neq y$ such that
$z_{T,c}(x+\delta)\geq z_{T,y}(x+\delta)$.
Therefore,
\begin{align}
\gamma(x)
&\leq
[z_{T,y}(x)-z_{T,c}(x)]
-
[z_{T,y}(x+\delta)-z_{T,c}(x+\delta)]
\notag\\
&\leq
2\|z_T(x)-z_T(x+\delta)\|_\infty
\label{eq:margin_lipschitz}\\
&\leq
2L\|\delta\|_\infty .
\notag
\end{align}

Combining this inequality with Eq.~\eqref{eq:ce_margin} yields

\begin{equation}
\|\delta\|_\infty
\geq
\frac{\gamma(x)}{2L}
\geq
\frac{
-\log\left(
e^{\mathrm{CE}(z_T(x),y)}-1
\right)
}{2L}.
\label{eq:robust_bound}
\end{equation}

When the right-hand side is positive, Eq.~\eqref{eq:robust_bound} provides a lower bound on the perturbation magnitude required to change the teacher prediction.
Under a common Lipschitz bound for the fixed robust teacher, reducing cross-entropy increases this lower bound on the perturbation required to change the teacher prediction.
This motivates searching for teacher-collaborative examples with lower cross-entropy to provide additional teacher supervision during distillation.

\subsection{Constrained Joint Inner Search}
\label{subsec:joint_inner_search}

Let $S_\theta$ and $T$ denote the student and fixed robust teacher, with logits
$z_S(x)$ and $z_T(x)$.
Their predictive distributions are
$p_S(x)=\operatorname{softmax}(z_S(x))$,
$p_T(x)=\operatorname{softmax}(z_T(x))$.
The perturbation neighborhood is defined as
\begin{equation}
\mathcal{B}_{\varepsilon}(x)
=
\{x' : \|x'-x\|_\infty\leq\varepsilon\}.
\end{equation}

For each training example $(x,y)$, CGARD searches for a student-adversarial example $x'_S$ and a teacher-collaborative example $x''_T$ within $\mathcal{B}_{\varepsilon}(x)$.
The two examples are treated as separate optimization variables, while their updates are jointly determined by the following discrepancy between teacher and student predictions:
\begin{equation}
\begin{gathered}
\max_{x'_S,x''_T\in\mathcal{B}_{\varepsilon}(x)}
\quad
D_{\mathrm{SKL}}
\left(
p_S(x'_S),p_T(x''_T)
\right)
\\
\mathrm{s.t.}\quad
\mathrm{CE}\left(z_T(x''_T),y\right)
\leq
\mathrm{CE}\left(z_T(x),y\right),
\end{gathered}
\label{eq:inner}
\end{equation}
where
\begin{equation}
D_{\mathrm{SKL}}(p,q)
=
\frac{1}{2}
\left[
\mathrm{KL}(p\|q)+\mathrm{KL}(q\|p)
\right].
\label{eq:skl}
\end{equation}

In particular, the update of $x''_T$ depends on the current student prediction at $x'_S$.
The constraint $\mathrm{CE}(z_T(x''_T),y)\leq \mathrm{CE}(z_T(x),y)$ requires the teacher to retain at least the same confidence in the ground-truth class at $x''_T$ as on the clean example $x$.

Both search examples are initialized from the $x$ with standard Gaussian perturbations scaled by 0.001.
We approximately optimize Eq.~\eqref{eq:inner} using iterative gradient ascent.
After each update, both examples are projected onto the perturbation neighborhood and clipped to the valid pixel range, while the teacher cross-entropy constraint is enforced by the feasibility check described below.
\begin{equation}
\begin{aligned}
x'^{\,k+1}_S
&=
\Pi_{\mathcal{B}_{\varepsilon}(x)}
\left(
x'^{\,k}_S
+
\eta\,
\operatorname{sign}
\nabla_{x'_S} D_{\mathrm{SKL}}
\right), \\
\widetilde{x}''^{\,k+1}_T
&=
\Pi_{\mathcal{B}_{\varepsilon}(x)}
\left(
x''^{\,k}_T
+
\eta\,
\operatorname{sign}
\nabla_{x''_T} D_{\mathrm{SKL}}
\right).
\end{aligned}
\label{eq:inner_update}
\end{equation}

The teacher parameters remain fixed throughout the search, and the gradients with respect to $x''_T$ are used only to update the teacher-collaborative example.
For each training example, the updated candidate in the teacher branch is retained only when
\begin{equation}
\mathrm{CE}
\left(
z_T(\widetilde{x}''^{\,k+1}_T),y
\right)
\leq
\mathrm{CE}(z_T(x),y).
\end{equation}

Otherwise, the candidate is reset to the clean example before the next update.

\subsection{Dual-Alignment Outer Minimization}
\label{subsec:outer_minimization}

Given the generated examples $x'_S$ and $x''_T$, CGARD keeps the teacher fixed and trains the student with a collaborative alignment term and an adversarial alignment term:
\begin{equation}
\begin{aligned}
\mathcal{L}_{\mathrm{col}}
&=
\mathrm{KL}
\left(
p_T(x''_T)\,\|\,p_S(x'_S)
\right), \\
\mathcal{L}_{\mathrm{adv}}
&=
\mathrm{KL}
\left(
p_T(x'_S)\,\|\,p_S(x'_S)
\right).
\end{aligned}
\label{eq:outer_terms}
\end{equation}

The collaborative alignment term $\mathcal{L}_{\mathrm{col}}$ aligns the student's prediction on $x'_S$ with the teacher's prediction on $x''_T$.
The adversarial alignment term $\mathcal{L}_{\mathrm{adv}}$ aligns the teacher and student predictions on the same student-adversarial example $x'_S$.
Thus, the two terms provide teacher supervision from different examples for the same student prediction on $x'_S$.
\begin{equation}
\mathcal{L}_{\mathrm{CGARD}}
=
(1-\alpha)\mathcal{L}_{\mathrm{col}}
+
\alpha\mathcal{L}_{\mathrm{adv}},
\label{eq:outer}
\end{equation}
Here, $\alpha\in[0,1]$ balances the two alignment terms.
At each training iteration, the constrained joint inner search regenerates $x'_S$ and $x''_T$, followed by minimization of $\mathcal{L}_{\mathrm{CGARD}}$ with respect to the student parameters.

\section{Experiments}
\label{Experiments}

\subsection{Experimental Setup}

We evaluate CGARD on CIFAR-10 and CIFAR-100~\cite{cifar100_2009}.
We use WideResNet-34-10 (WRN)~\cite{wrnarxiv2016} pretrained with ST~\cite{stat2023iclr} as the fixed robust teacher.
ResNet-18 (RN-18)~\cite{resnet18cvpr2016} and MobileNet-V2 (MN-V2)~\cite{mnv2cvpr2018} are used as students.
We compare CGARD with PGD-AT~\cite{atpgd_ICLRmadry2018}, TRADES~\cite{at_tradesicml2019}, ST~\cite{stat2023iclr}, ARD~\cite{AD_ard2020aaai}, RSLAD~\cite{AD_rslad2021iccv}, IAD~\cite{AD_iad2022iclr}, AdaAD~\cite{AdaAD2023cpvr}, and PeerAiD~\cite{AD_peeraid2024cvpr}. 
All baseline results are obtained from our reproductions.

We report clean, PGD~\cite{atpgd_ICLRmadry2018}, CW$_\infty$~\cite{att_cwsp2017_qianyi}, and standard AutoAttack (AA)~\cite{att_aaicml20} accuracy, with FGSM~\cite{fgsmiclr2015} additionally used for ablation studies.
All attacks use an $\ell_\infty$ perturbation budget of $\varepsilon=8/255$; PGD and CW$_\infty$ use 20 iterations with a step size of $2/255$.

CGARD students are trained for 300 epochs using SGD with momentum 0.9, weight decay $2\times10^{-4}$, and a batch size of 128.
The initial learning rate is 0.1 and is divided by 10 at epochs 215, 260, and 285.
The inner search uses $K=10$ iterations with step size $2/255$, and $\alpha=0.7$.

\subsection{White-Box Robustness}
\label{subsec:whitebox}

\begin{table*}[t]
\small
\centering
\renewcommand{\arraystretch}{0.98}
\setlength{\tabcolsep}{1.4pt}

\caption{White-box robustness comparison (\%) on CIFAR-10 and CIFAR-100 under
$\ell_\infty$ attacks with $\varepsilon=8/255$.
Best and second-best student results are shown in bold and underlined, respectively.}
\label{tab:main_whitebox}

\begin{tabular*}{\textwidth}
{@{\extracolsep{\fill}}cllcccccccc@{}}
\hline

&
&
&
\multicolumn{4}{c}{\textbf{CIFAR-10}}
&
\multicolumn{4}{c}{\textbf{CIFAR-100}}
\\

\cline{4-7}
\cline{8-11}

\textbf{Type}
& \textbf{Arch.}
& \textbf{Method}
& \textbf{Clean}
& \textbf{PGD}
& \textbf{CW$_\infty$}
& \textbf{AA}
& \textbf{Clean}
& \textbf{PGD}
& \textbf{CW$_\infty$}
& \textbf{AA}
\\

\hline

\textbf{Teacher}
& WRN
& ST~\cite{stat2023iclr}
& 84.75
& 57.83
& 55.49
& 54.33
& 59.40
& 33.71
& 30.30
& 29.19
\\

\hline

\multirow[c]{20}{*}{\textbf{Student}}
& \multirow[c]{10}{*}{RN-18}
& Natural
& \textbf{95.28}
& 0.01
& 0.01
& 0.00
& \textbf{77.72}
& 0.00
& 0.00
& 0.00
\\

&
&
PGD-AT~\cite{atpgd_ICLRmadry2018}
& 82.50
& 51.53
& 50.57
& 48.04
& 57.52
& 28.50
& 27.12
& 24.68
\\

&
&
TRADES~\cite{at_tradesicml2019}
& 82.78
& 50.83
& 49.37
& 47.63
& 56.32
& 26.95
& 23.77
& 22.69
\\

&
&
ST~\cite{stat2023iclr}
& 83.43
& 53.72
& 51.66
& 50.01
& 55.75
& 30.92
& 26.89
& 26.04
\\

&
&
ARD~\cite{AD_ard2020aaai}
& 83.29
& 53.65
& 51.63
& 50.01
& 58.44
& 31.54
& 28.47
& 26.42
\\

&
&
RSLAD~\cite{AD_rslad2021iccv}
& 82.98
& 55.61
& 53.00
& 51.49
& 56.55
& 32.85
& 29.34
& \underline{28.00}
\\

&
&
IAD~\cite{AD_iad2022iclr}
& 82.84
& 54.13
& 51.72
& 50.10
& 57.82
& 31.94
& 28.32
& 26.80
\\

&
&
AdaAD~\cite{AdaAD2023cpvr}
& \underline{84.28}
& \underline{56.17}
& \underline{53.36}
& \underline{52.02}
& \underline{58.91}
& 32.57
& 29.00
& 27.80
\\

&
&
PeerAiD~\cite{AD_peeraid2024cvpr}
& 82.63
& 54.21
& 52.24
& 50.25
& 56.76
& \underline{32.99}
& \underline{29.66}
& 27.97
\\

&
&
\textbf{CGARD}
& 82.96
& \textbf{56.47}
& \textbf{53.69}
& \textbf{52.46}
& 56.93
& \textbf{33.28}
& \textbf{29.83}
& \textbf{28.49}
\\

\cline{2-11}

&
\multirow[c]{10}{*}{MN-V2}
& Natural
& \textbf{91.82}
& 0.00
& 0.00
& 0.00
& \textbf{71.92}
& 0.00
& 0.00
& 0.00
\\

&
&
PGD-AT~\cite{atpgd_ICLRmadry2018}
& 79.99
& 49.23
& 47.70
& 45.11
& 56.00
& 27.73
& 25.84
& 23.25
\\

&
&
TRADES~\cite{at_tradesicml2019}
& 80.98
& 49.97
& 47.76
& 46.51
& 56.82
& 28.22
& 24.16
& 23.18
\\

&
&
ST~\cite{stat2023iclr}
& 80.05
& 51.97
& 48.14
& 47.12
& 54.53
& 30.10
& 25.41
& 24.64
\\

&
&
ARD~\cite{AD_ard2020aaai}
& 81.80
& 52.50
& 50.22
& 48.67
& 57.25
& 31.00
& 27.90
& 26.02
\\

&
&
RSLAD~\cite{AD_rslad2021iccv}
& 83.11
& \underline{54.24}
& \underline{51.68}
& \underline{50.28}
& 56.78
& \underline{32.11}
& 28.43
& \underline{26.94}
\\

&
&
IAD~\cite{AD_iad2022iclr}
& 81.44
& 52.47
& 50.15
& 48.06
& 57.41
& 31.59
& 28.09
& 26.36
\\

&
&
AdaAD~\cite{AdaAD2023cpvr}
& \underline{83.86}
& 54.06
& 51.41
& 49.95
& \underline{57.95}
& 30.92
& 27.00
& 25.55
\\

&
&
PeerAiD~\cite{AD_peeraid2024cvpr}
& 82.12
& 53.70
& 50.76
& 48.76
& 56.67
& 31.47
& \underline{28.82}
& 26.92
\\

&
&
\textbf{CGARD}
& 82.19
& \textbf{54.96}
& \textbf{52.12}
& \textbf{50.86}
& 56.79
& \textbf{32.95}
& \textbf{29.00}
& \textbf{27.66}
\\

\hline
\end{tabular*}
\end{table*}

Table~\ref{tab:main_whitebox} reports clean and white-box robust accuracy on CIFAR-10 and CIFAR-100.
CGARD achieves the highest PGD, CW$_\infty$, and AA accuracy among the evaluated student methods in all four dataset--architecture settings.
On CIFAR-10, its AA accuracies are 52.46\% for RN-18 and 50.86\% for MN-V2, exceeding the strongest corresponding AA baselines by 0.44 and 0.58 percentage points.
On CIFAR-100, the corresponding accuracies are 28.49\% and 27.66\%, with gains of 0.49 and 0.72 percentage points.
Compared to these strongest AA baselines, the clean accuracy decreases by 1.32 and 0.92 percentage points on CIFAR-10, respectively, while it is maintained or improved on CIFAR-100.
These results indicate consistent robustness gains across the evaluated settings, while changes in clean accuracy vary across datasets.

\subsection{Black-Box Transfer Robustness}
\label{subsec:blackbox}

\begin{table}[t]
\small
\centering
\renewcommand{\arraystretch}{0.98}
\setlength{\tabcolsep}{2.0pt}

\caption{Black-box transfer robustness (\%) on CIFAR-10 with robust WRN, VGG-19, and MN-V2 surrogates and an RN-18 target.
Best and second-best results are shown in bold and underlined.}
\label{tab:blackbox}

\begin{tabular}{lcccccc}
\hline

\multicolumn{1}{c}{\textbf{Surrogate Model}}
& \multicolumn{2}{c}{\textbf{WRN}}
& \multicolumn{2}{c}{\textbf{VGG-19}}
& \multicolumn{2}{c}{\textbf{MN-V2}} \\

\cmidrule(lr){2-3}
\cmidrule(lr){4-5}
\cmidrule(lr){6-7}

\textbf{Method}
& \textbf{PGD}
& \textbf{CW$_\infty$}
& \textbf{PGD}
& \textbf{CW$_\infty$}
& \textbf{PGD}
& \textbf{CW$_\infty$} \\

\hline

PGD-AT~\cite{atpgd_ICLRmadry2018}
& 65.21
& 65.21
& 63.84
& 63.12
& 63.25
& 62.41 \\

TRADES~\cite{at_tradesicml2019}
& 65.64
& 65.26
& 63.89
& 62.96
& 63.00
& 62.26 \\

ST~\cite{stat2023iclr}
& \underline{66.92}
& 66.54
& 66.41
& 65.70
& 66.10
& \underline{65.49} \\

ARD~\cite{AD_ard2020aaai}
& 66.25
& 66.06
& 65.57
& 64.76
& 64.85
& 64.19 \\

RSLAD~\cite{AD_rslad2021iccv}
& 66.70
& \underline{66.66}
& 66.28
& 65.69
& 65.82
& 65.16 \\

IAD~\cite{AD_iad2022iclr}
& 66.59
& 66.33
& 65.86
& 64.95
& 65.31
& 64.52 \\

AdaAD~\cite{AdaAD2023cpvr}
& 66.80
& 66.53
& \underline{66.59}
& \underline{65.88}
& \underline{66.15}
& 65.42 \\

PeerAiD~\cite{AD_peeraid2024cvpr}
& 66.64
& 66.62
& 66.19
& 65.51
& 65.03
& 64.65 \\

\textbf{CGARD}
& \textbf{66.94}
& \textbf{66.69}
& \textbf{66.60}
& \textbf{66.06}
& \textbf{66.19}
& \textbf{65.52} \\

\hline
\end{tabular}
\end{table}

We evaluate transfer-based black-box robustness on CIFAR-10 using WRN, VGG-19, and MN-V2 as surrogate architectures and RN-18 as the target student architecture.
Table~\ref{tab:blackbox} reports robust accuracy against transferred PGD and CW$_\infty$ attacks.
CGARD shows competitive transfer robustness across the evaluated surrogate models and attacks.
These results complement the white-box evaluation across different surrogate architectures.

\subsection{Ablation Studies}
\label{subsec:ablation}

\begin{table}[t]
\small
\centering
\renewcommand{\arraystretch}{1.0}
\setlength{\tabcolsep}{3.2pt}

\caption{Comparison of sequential search and constrained joint inner search on CIFAR-10 with WRN as the teacher and RN-18 as the student.
Best results are shown in bold.}
\label{tab:ablation_jis}

\begin{tabular}{cccccc}
\hline
\textbf{Configuration}
& \textbf{Clean}
& \textbf{FGSM}
& \textbf{PGD}
& \textbf{CW$_\infty$}
& \textbf{AA} \\
\hline

Sequential Search
& \textbf{84.25}
& 60.97
& 54.92
& 52.36
& 51.01 \\

Joint Search \textbf{(CGARD)}
& 82.96
& \textbf{61.12}
& \textbf{56.47}
& \textbf{53.69}
& \textbf{52.46} \\

\hline
\end{tabular}
\end{table}

\begin{table}[t]
\small
\centering
\renewcommand{\arraystretch}{1.0}
\setlength{\tabcolsep}{3.5pt}

\caption{Ablation study of the dual-alignment objective on CIFAR-10 with WRN as the teacher and RN-18 as the student. Best results are shown in bold.}
\label{tab:ablation_outer}

\begin{tabular}{ccccccc}
\hline
$\boldsymbol{\mathcal{L}_{\mathrm{col}}}$
& $\boldsymbol{\mathcal{L}_{\mathrm{adv}}}$
& \textbf{Clean}
& \textbf{FGSM}
& \textbf{PGD}
& \textbf{CW$_\infty$}
& \textbf{AA} \\
\hline

$\checkmark$
&
& 82.64
& 60.65
& 56.06
& 53.21
& 51.84 \\

&
$\checkmark$
& \textbf{83.13}
& 60.71
& 55.95
& 53.07
& 51.83 \\

$\checkmark$
& $\checkmark$
& 82.96
& \textbf{61.12}
& \textbf{56.47}
& \textbf{53.69}
& \textbf{52.46} \\

\hline
\end{tabular}
\end{table}

Table~\ref{tab:ablation_jis} compares two strategies for example generation.
The sequential strategy first follows ST~\cite{stat2023iclr} to generate both teacher-collaborative and teacher-adversarial examples, fixes the teacher prediction on $x''_T$, and then generates $x'_S$ by maximizing its KL divergence from the fixed teacher prediction.
CGARD instead jointly optimizes $x'_S$ and $x''_T$ through $D_{\mathrm{SKL}}$ under the teacher cross-entropy constraint.
Compared to the sequential strategy, the proposed joint search strategy improves PGD, CW$_\infty$, and AA accuracy by 1.55, 1.33, and 1.45 percentage points, respectively, while clean accuracy decreases by 1.29 percentage points.

Table~\ref{tab:ablation_outer} keeps the constrained joint inner search fixed.
Using $\mathcal{L}_{\mathrm{col}}$ or $\mathcal{L}_{\mathrm{adv}}$ alone yields AA accuracies of 51.84\% and 51.83\%, compared with 52.46\% when both terms are used.

\section{Conclusion}
\label{Conclusion}

We proposed CGARD, which jointly optimizes separate student-adversarial and teacher-collaborative examples under a teacher cross-entropy constraint and trains the student through dual alignment.
Across CIFAR-10 and CIFAR-100 with two student architectures, CGARD achieves the highest AA accuracy among the evaluated student methods and competitive black-box transfer robustness.
The comparison of search strategies and the ablations of the outer objective support the constrained joint search and the combined use of the two alignment terms, respectively.

\section{Acknowledgment}
This work is supported by the Yunnan Research Project (Grant Nos. 202505AF350053, 202503AG380006, 202401AT- 070474, 202501AU070059, 202403AP140021, and 202601CI- 070131), and the National Natural Science Foundation of China (Grant Nos. 62562061, 62502422, and 62462067).

\bibliographystyle{IEEEbib}
\bibliography{ref}

\end{document}